\documentclass{llncs}
\usepackage{graphicx}
\usepackage{epstopdf}
\usepackage{subfigure}
\usepackage{multirow}
\usepackage{amsmath}
\usepackage{url}
\usepackage{tikz}
\usepackage{balance}
\usepackage{hyperref}
\usepackage{adjustbox,lipsum}
\usepackage[ruled,vlined]{algorithm2e}
\usepackage[english]{babel}
\usepackage[utf8x]{inputenc}
\usepackage[T1]{fontenc}
\usepackage{float}
\usepackage{afterpage}

\begin{document}
\newcommand*{\textred}{\textcolor{red}}
\newcommand*{\textblue}{\textcolor{blue}}
\newcommand*{\textgreen}{\textcolor{green}}

\title{High frame-rate cardiac ultrasound imaging with deep learning}
%\title{MLA2SLA: Ultrasound SLA image quality from MLA using Deep Learning}
\author{Ortal Senouf\inst{1} \and Sanketh Vedula\inst{1} \and Grigoriy Zurakhov\inst{1} \and Alex Bronstein\inst{1}  \and \\ Michael Zibulevsky\inst{1} \and Oleg Michailovich\inst{2} \and  Dan Adam\inst{1} \and David Blondheim\inst{3}}

 \institute{Technion - Israel Institute of Technology \\ \email{\{senouf,sanketh\}@campus.technion.ac.il}\and University of Waterloo, Canada \and Hillel Yaffe Medical Center}

\maketitle              % typeset the title of the contribution

\begin{abstract}
Cardiac ultrasound imaging requires a high frame rate in order to capture rapid motion. This can be achieved by multi-line acquisition (MLA), where several narrow-focused received lines are obtained from each wide-focused transmitted line. This shortens the acquisition time at the expense of introducing block artifacts.  In this paper, we propose a data-driven learning-based approach to improve the MLA image quality. We train an end-to-end convolutional neural network on pairs of real ultrasound cardiac data, acquired through MLA and the corresponding single-line acquisition (SLA).  The network achieves a significant improvement in image quality for both $5-$ and $7-$line MLA resulting in a decorrelation measure similar to that of SLA while having the frame rate of MLA.
\end{abstract}
\section{Introduction}
Increasing the frame rate is a major challenge in 2D and 3D echocardiography. Investigating deformations at different stages of the cardiac cycle is crucial for cardiovascular imaging; hence high temporal resolution is highly desired in addition to the spatial resolution.  There are several ways to increase the frame rate of ultrasound imaging; one of the most commonly used techniques, which is implemented in many ultrasound scanners, is multi-line acquisition (MLA) \cite{PRB1984}, often referred to as parallel receive beamforming (PRB) \cite{hergum2007parallel}.
%The basic idea is to receive multiple pulses in receive for each transmit pulse sent; and thereby increase the frame rate by requiring less number of transmit in reconstructing the image. 

\paragraph{Single- vs. multi-line acquisition.}  In single-line acquisition (SLA), a narrow-focused pulse is transmitted by introducing transmit time delays through a linear phased array of acoustic transducer elements. Upon reception the obtained signal is dynamically focused along the receive (Rx) direction which is identical to the transmit (Tx) direction. The spatial region of interest is raster scanned line-by-line to obtain an ultrasound image. 

The need to transmit a large number of pulses sequentially results in a low frame rate and renders SLA inadequate for cardiovascular imaging, where a high frame rate is mandatory, especially for quantitative analysis or during stress tests. For the same reason, SLA is neither useful for scanning large fields of view in real time 3D imaging applications. 

\begin{figure}[th]
\centering
\includegraphics[width=\textwidth,height=0.4\textheight,keepaspectratio]{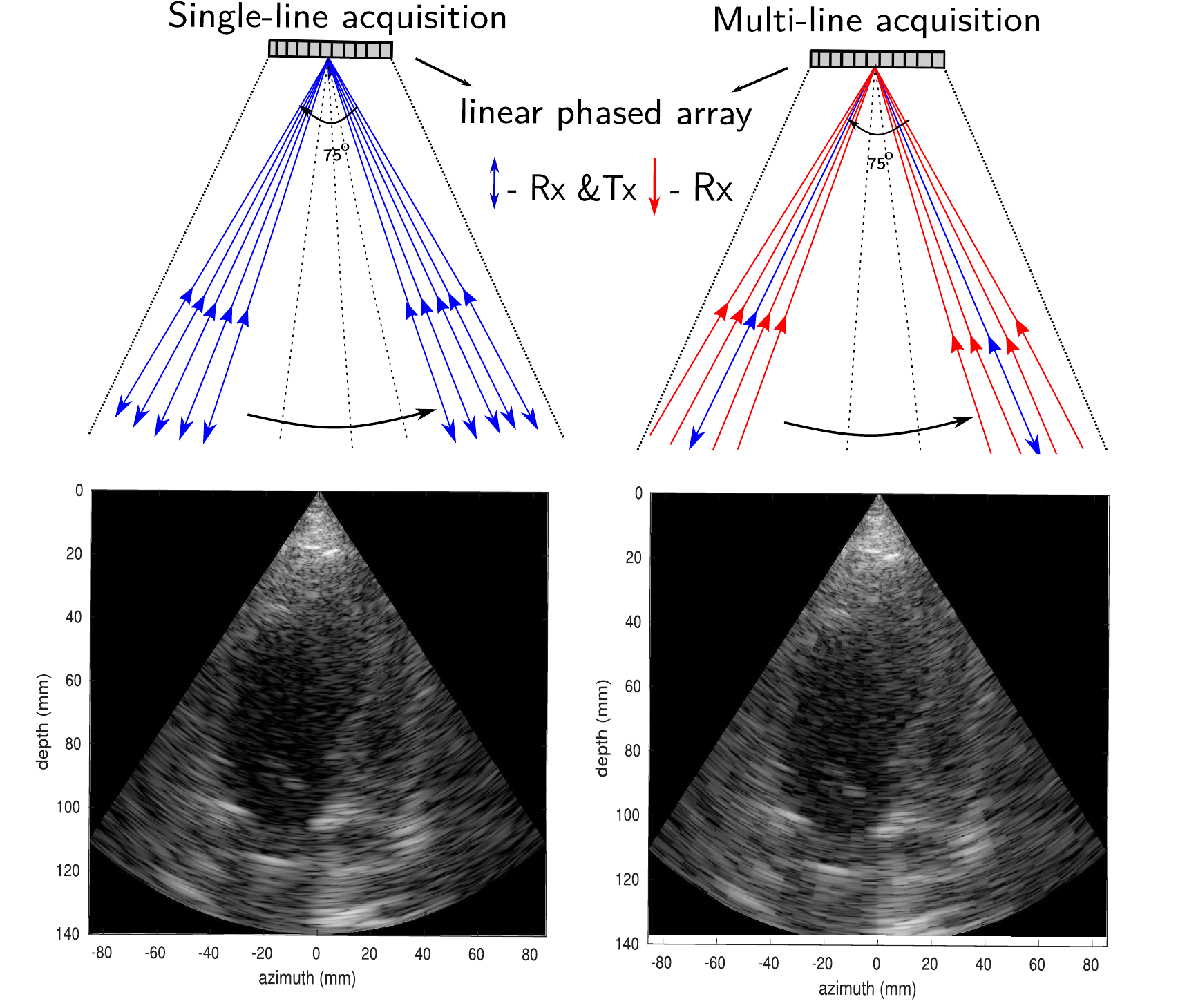}
\caption{Single (left) and multi-line (right, with MLA factor $m=5$) acquisition procedures and their corresponding ultrasound scans. Block artifacts can be seen along the axial direction in MLA. Zooming in is recommended.}
\label{MLASLA}
\end{figure}

In an attempt to overcome the frame rate problem, the MLA method was proposed in  \cite{PRB1984}, \cite{PRB1991}. The main idea behind MLA is to transmit a weakly focused beam
that provides a sufficiently wide coverage for a high number of received lines. On the receiver side, $m$ lines is constructed from the data acquired from each transmit event, thereby increasing the frame rate by $m$ (the latter number is usually referred to as the \emph{MLA factor}). Signal formation in the SLA and MLA modalities is demonstrated in Figure \ref{MLASLA} where $5$-MLA is depicted. For a $5$-MLA, we construct $5$ Rx lines per each Tx thus increasing the frame rate by the factor of $5$. 

\paragraph{MLA Artifacts.} As the Tx and Rx are no longer aligned in the MLA mode, the two-way beam profile is shifted towards the original transmit direction, making the lateral sampling irregular \cite{hergum2007parallel}. This \emph{beam warping} effect causes sharp lateral discontinuities that are manifested as block artifacts in the image domain. 

 The observed block artifacts in the ultrasound images (see, e.g., Figure \ref{MLASLA}) tend to be more obvious when the number of transmit events decreases. The MLA artifact can be measured by assessing the correlation coefficient between each two adjacent Rx lines in the in-phase and quadrature (I/Q) demodulated beamformed data \cite{bjastad2007impact}. In SLA or compensated MLA, the averaged correlation values inside MLA groups and between MLA groups are almost the same. In the uncompensated cases, however, the correlation values are different. 
 
 Apart from  beam warping, there are two other effects caused by the transmit-receive misalignment: \emph{skewing}, where shape of the two-way beam profile becomes asymmetric, and \emph{gain variation}, where the outermost lines inside the group have a lower gain than the innermost lines \cite{bjastad2007impact}. 

\paragraph{Related work.} Several methods have been proposed in literature to decrease MLA artifacts, including transmit sinc apodization \cite{augustine1987high} and dynamic steering \cite{thiele1994method}, incoherent interpolation \cite{holley1998ultrasound},\cite{liu2002system} (applied after envelope detection), and its coherent (before envelope detection) counterparts \cite{Wright1997STB},\cite{hergum2007parallel}.
 One of the more prominent methods, synthetic transmit beamforming (STB)\cite{hergum2007parallel}, creates synthetic Tx lines by coherently interpolating information received from each two adjacent Tx events in intermediate directions. This technique creates highly correlated lines, attenuating block artifacts. A common practice for MLA imaging with focused beams is to create $2-4$ Rx lines per each Tx event in cases without overlap, or $4-8$ lines in the presence of overlaps from adjacent transmissions, in order to perform the correction \cite{hergum2007parallel},\cite{bjastad2007impact},\cite{rabinovich2013multi}. Thus, creating eight lines with overlaps provides an effective frame rate increase by the factor of $4$. In this paper, however, we used odd MLA factors $m=5,7$ for the purpose of %training the network on %
 acquiring data from aligned directions for both SLA and MLA.

Recently, data-driven learning techniques based on convolutional neural networks (CNNs) have been extensively used for solving inverse problems in imaging and in medical imaging in particular, for example, in X-ray CT reconstruction and denoising \cite{Unser2017SPM} and in real-time ultrasound post-processing \cite{vedula2017towards}. Inspired by their success, we propose a data-driven approach to overcome MLA artifacts.

\paragraph{Contributions.} We propose an end-to-end CNN-based approach for MLA artifact correction. Our fully convolutional network consists of interpolation layers followed by a trainable apodization layer, and is trained on in-vivo cardiac data to approximate an SLA quality image. We demonstrate the effectiveness of this network both visually and quantitatively using the decorrelation measure ($D_c$) and SSIM \cite{wang2004image} quality criteria. To the best of our knowledge, this is the first study to report good artifact corrections in the case of $5-$ $7-$MLA. We show that the trained network generalizes well across patients, as well as to phantom data.

\section{Methods}
\subsection{Improving MLA with CNNs}\label{sub:MLACNN}     
\begin{figure}[h]
\includegraphics[width=\textwidth,keepaspectratio]{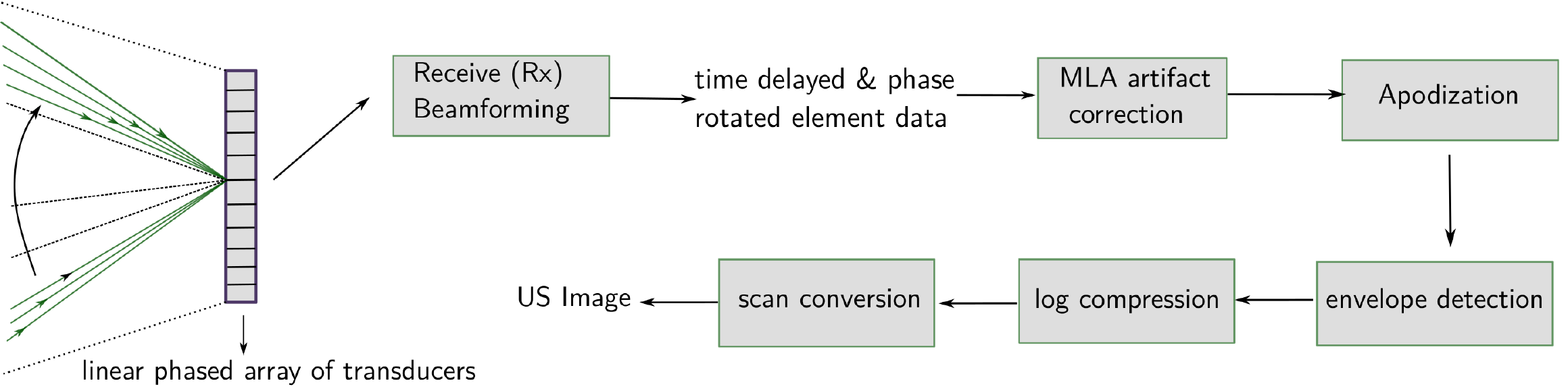}
\caption{Traditional MLA ultrasound imaging pipeline.}
\label{TraditionalMLA}
\end{figure}

\begin{figure}[h]
\centering
\includegraphics[width=\textwidth]{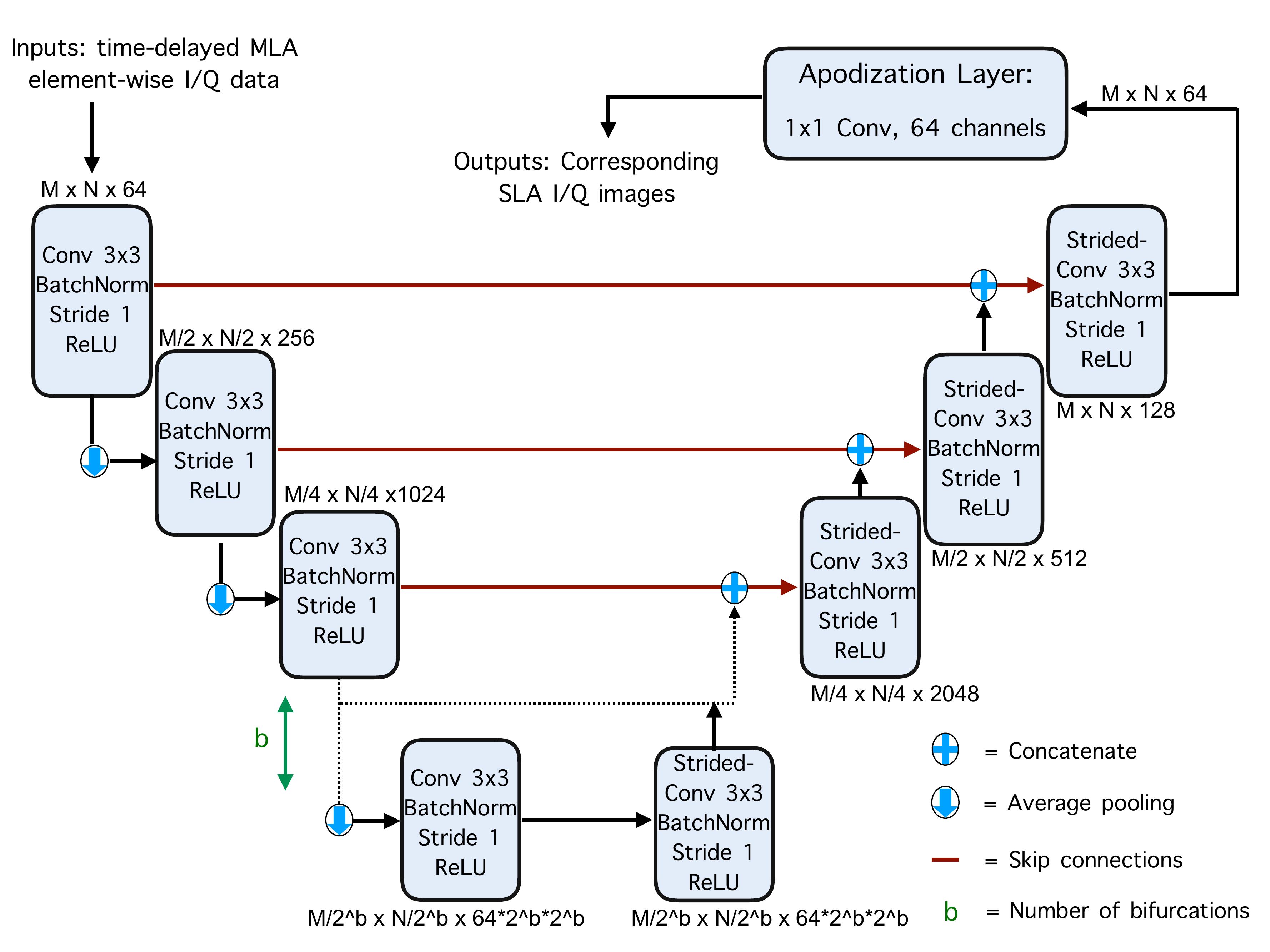}
\caption{Proposed CNN-based MLA artifact correction pipeline.}
\label{NetArch}
\end{figure}

Aiming at providing a general and optimal solution for MLA interpolation achieving SLA quality, we propose to replace MLA artifact correction and apodization phases in the traditional MLA pipeline as shown in Figure \ref{TraditionalMLA} with an end-to-end CNN depicted in Figure \ref{NetArch}. We draw similarities to \cite{rabinovich2013multi} who showed that combining MLA interpolation with an optimal apodization method produces superior results compared to the traditional approaches. 
Our network comprises both the interpolation and the apodization stages that are trained jointly. 

\paragraph{Interpolation stage.} The interpolation stage consists of our CNN containing $10$ convolutional layers with symmetric skip connections\cite{mao2016image},\cite{ronneberger2015u} from each layer in the downsampling track to its corresponding layer in the upsampling track as visualized in Figure \ref{NetArch}. Downsampling is performed using average pooling and strided convolutions are used for upsampling. The number of bifurcations is set to $5$ for all the experiments. The interpolation stage takes as an input the time-delayed and phase-rotated element-wise I/Q data from the transducer.   

\paragraph{Apodization stage.} Following the interpolation stage, we introduce a convolutional layer to perform apodization. This is performed using point-wise convolutions ($1\times1$) for each element's channel in the network and the results are then added to the learned weights of the convolution. The weights of the channel are initialized with a Hann window. 

\paragraph{Optimization.} We use the $L_1$ norm training loss to measure the discrepancy between the image predicted by the network and the ground truth SLA images. The loss is minimized using the Adam optimizer \cite{kingma2014adam} with a learning rate of $10^{-4}$. We observed that adding the apodization stage accelerates the training process, and makes the network converge faster. 

\subsection{Data acquisition and training}\label{sub:training}
We generated a dataset for training the network using cardiac data from six patients; each patient contributed $4$-$5$ cine loops, containing $32$ frames. The data was acquired using a GE experimental breadboard ultrasound system. The same transducer was used for both phantom and cardiac acquisition.  Excitation sinusoidal pulses of $1.75$ cycles, centered around $2.5$ MHz, were transmitted using $28$ central elements out of the total $64$ element in the probe with a pitch of $0.3$ mm, elevation size of $13$ mm and elevation focus of $100$ mm. The depth focus was set at $71$ mm. In order to assess the desired aperture for MLA setup, Field II simulator \cite{jensen1996field} was used as in \cite{rabinovich2013multi} using the transducer impulse response and tri-state transmission excitation sequence, requiring a minimal insonification of $-3$dB for all MLAs from a single Tx. 

On the Rx side, the I/Q demodulated signals were dynamically focused using linear interpolation, with an 	f-number of $1$. The FOV was covered with $140/140$ Tx/ Rx lines in SLA mode, $28/140$ Tx/Rx lines in the $5-$MLA mode, and $20/140$ Tx/Rx lines in the $7-$MLA mode. For both phantom and cardiac cases, the data were acquired in the SLA mode; $5-$MLA and $7-$MLA data was obtained by appropriately decimating the Rx pre-beamformed data.  

In total, we used $745$ frames from five patients for training and validation, while keeping the cine loops from the sixth patient for testing. The data set  comprised pairs of beamformed I/Q images with Hann window apodization, and the corresponding $5-$ and $7-$MLA pre-apodization samples with the dimensions of $652 \times 64 \times 140$ (depth $\times$ elements $\times$ Rx lines). The MLA data was acquired by decimation of the Tx lines of the SLA samples by the MLA factor ($m=5,7$).

We  trained dedicated CNNs for the reconstruction of SLA images from $5-$ and $7-$MLA. Each CNN was trained to a maximum of $200$ epochs on mini batches of size $4$.

\section{Experimental evaluation}
\subsection{Settings}
In order to assess the performance of our trained networks, we used cine loops from one patient excluded from the training/validation set. From two cine loops, each containing $32$ frames, we generated pairs of $5-$ and $7-$MLA samples and their corresponding SLA images the same way as described in section \ref{sub:training}, resulting in $64$ test samples. For quantitative evaluation of the performance of our method we measured the decorrelation ($D_c$) criterion that evaluates the artifact strength \cite{bjastad2007impact}, and the SSIM \cite{wang2004image} structural similarity criterion with respect to the SLA image. In addition, we tested the performance of our networks on four frames acquired from the GAMMEX Ultrasound 403GS LE Grey Scale Precision Phantom.        

\subsection{Results}
\begin{figure}[h]
\begin{minipage}[]{\linewidth}
	\begin{tabular}{ c@{\hskip 0.001\textwidth}c@{\hskip 0.001\textwidth}c@{\hskip 0.001\textwidth}c} 

		\includegraphics[width = 0.33\textwidth]{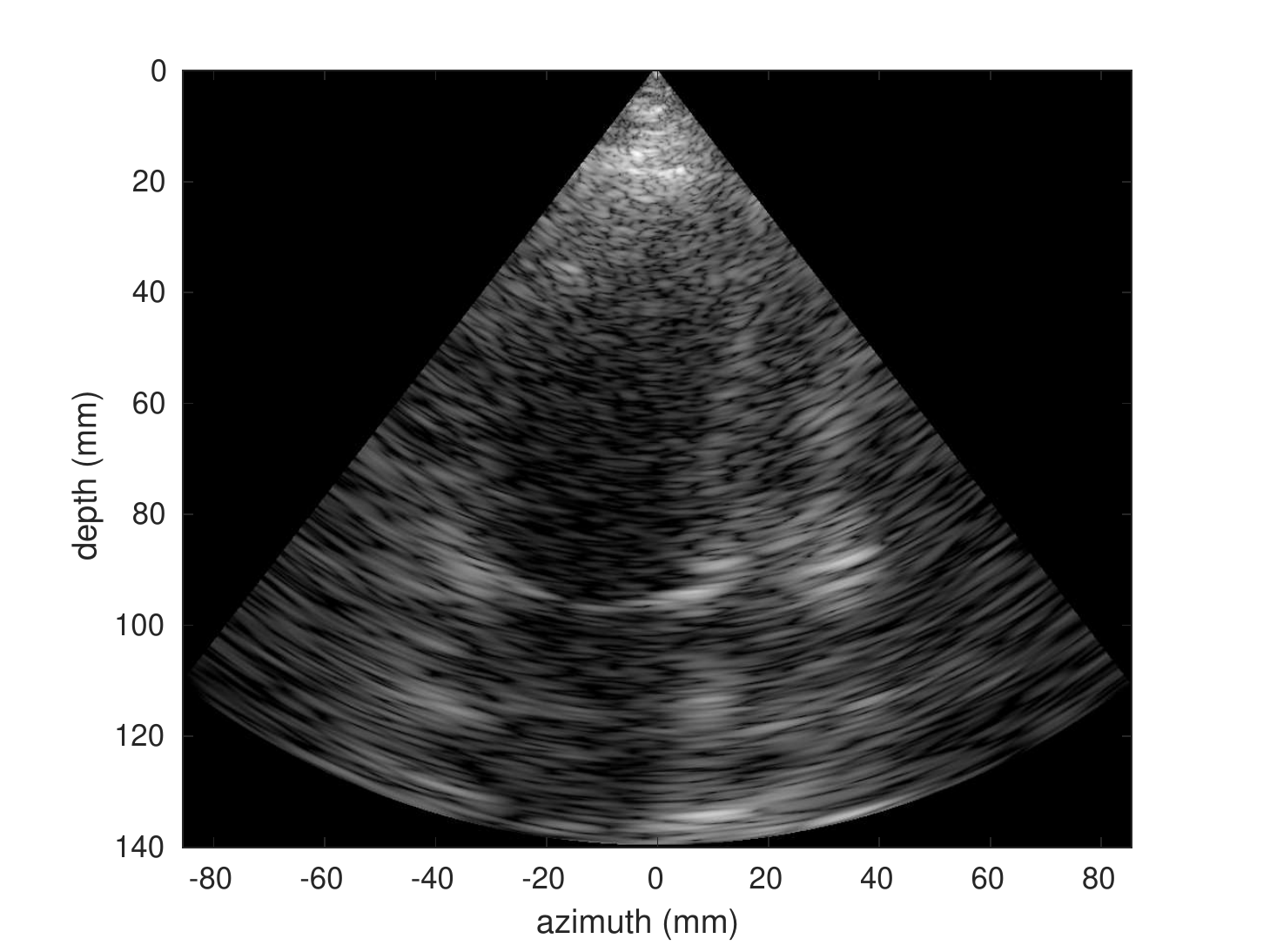} &
		\includegraphics[width = 0.33\textwidth]{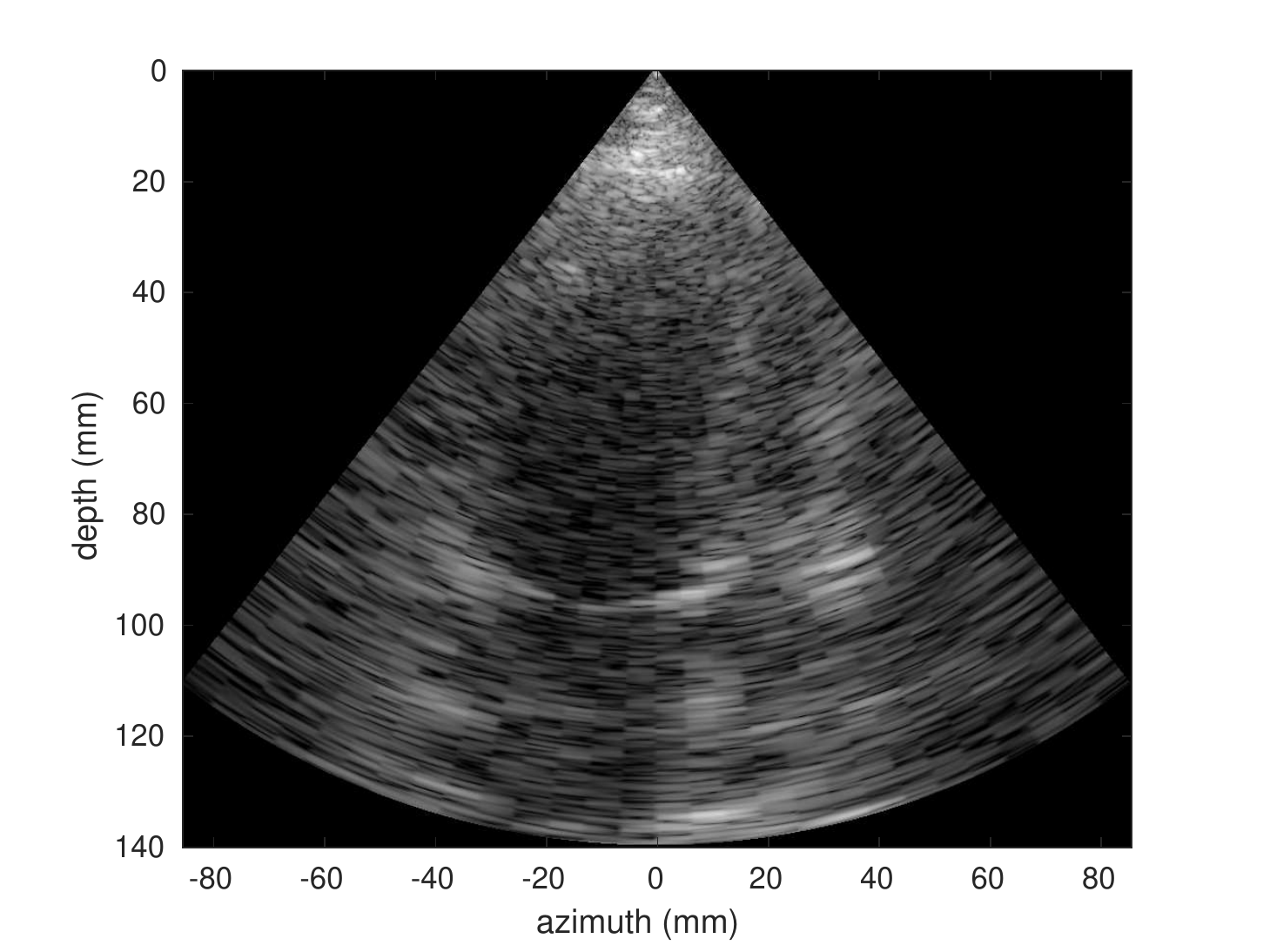} &
		\includegraphics[width = 0.33\textwidth]{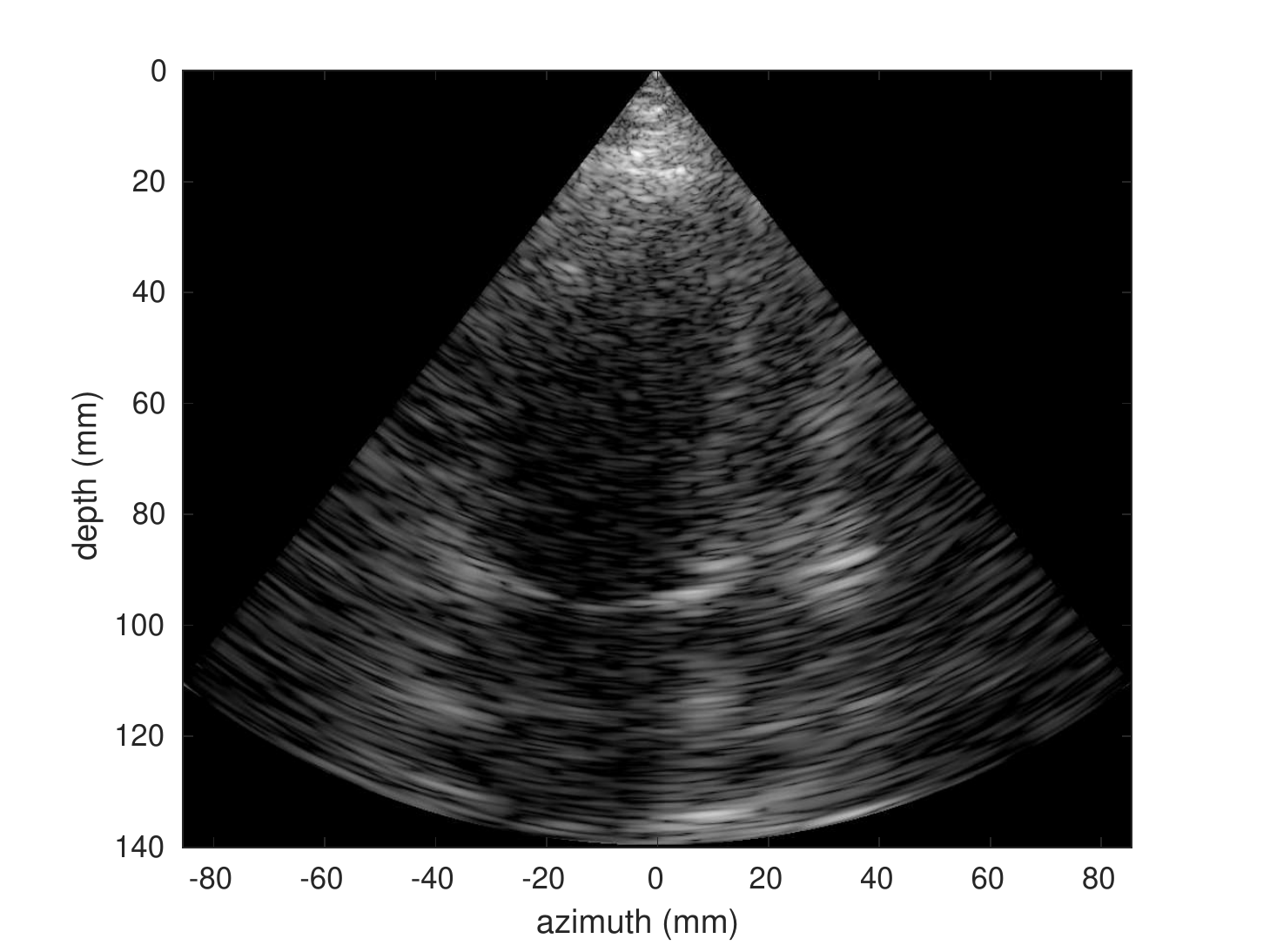} &
	       \\
        \includegraphics[width = 0.33\textwidth]{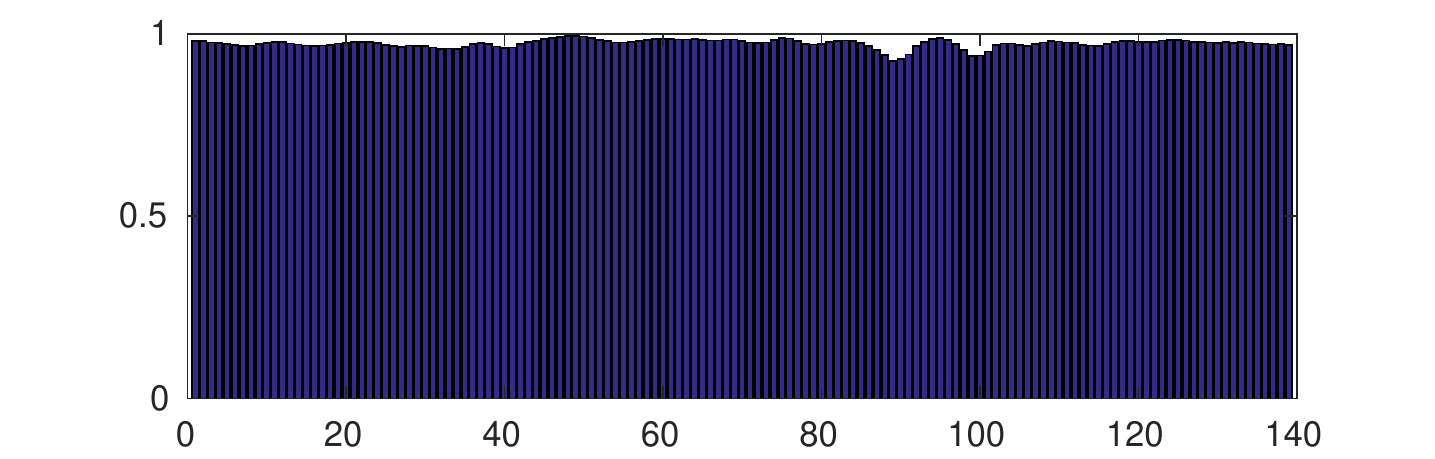} &
		\includegraphics[width = 0.33\textwidth]{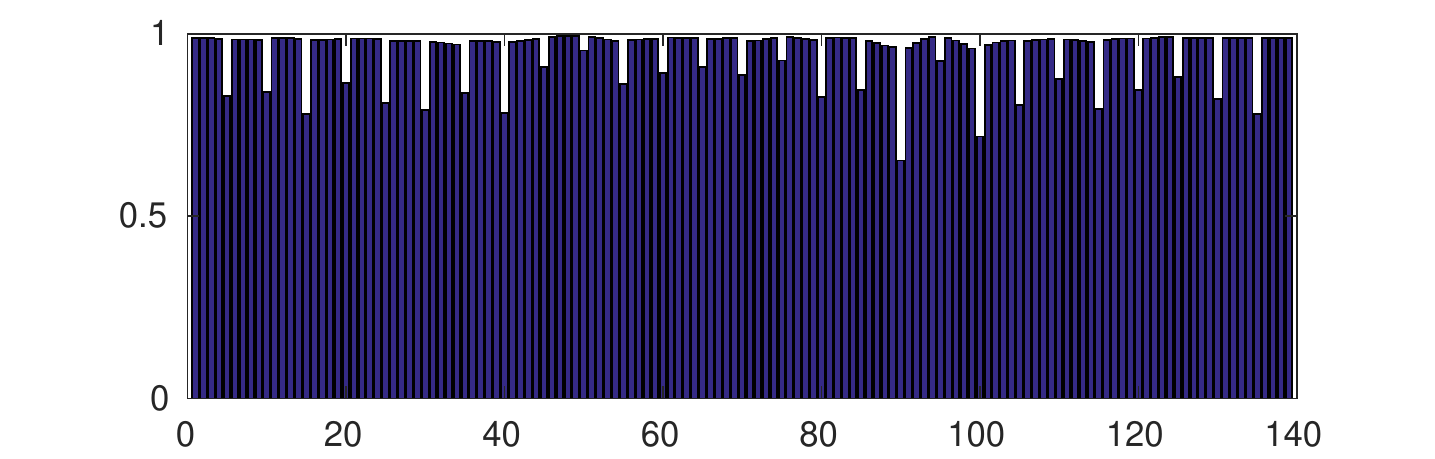} &
		\includegraphics[width = 0.33\textwidth]{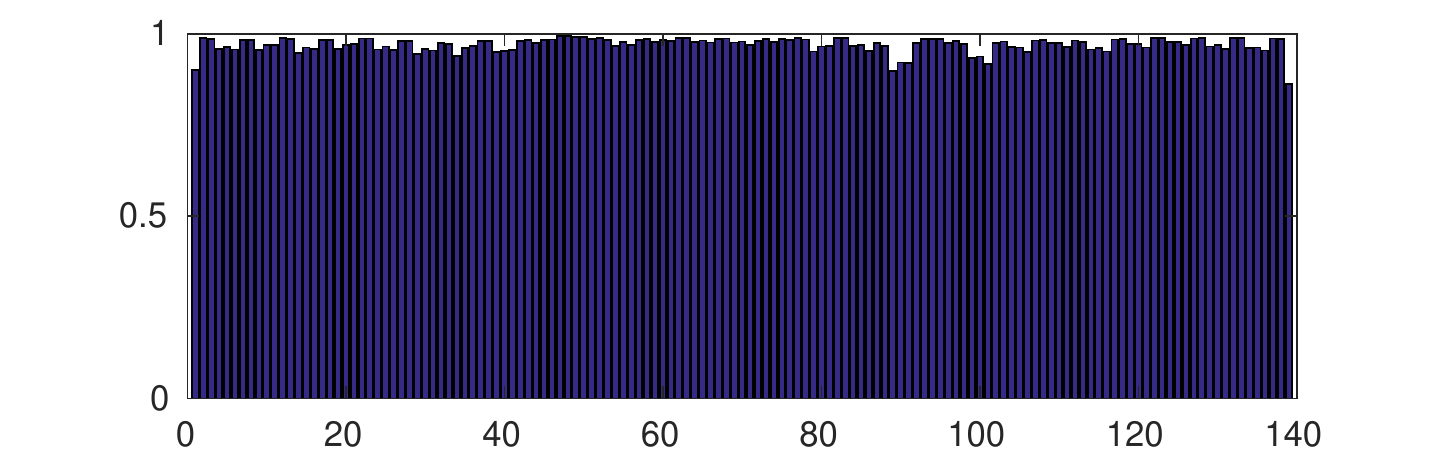}&
         \\
               (a) SLA & (b) $5-$MLA & (c) Corrected $5-$MLA & \\
           &
		\includegraphics[width = 0.33\textwidth]{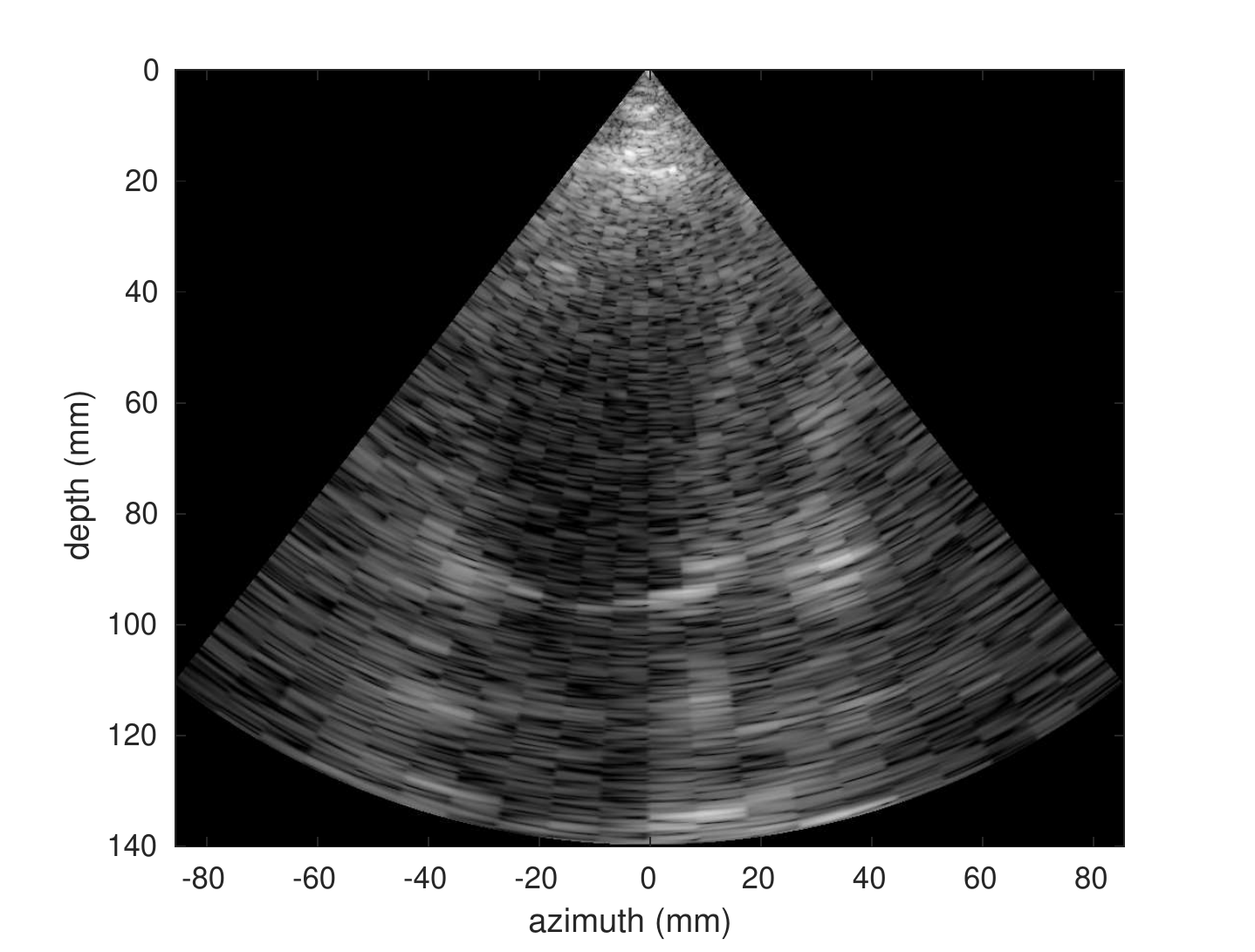} &
		\includegraphics[width = 0.33\textwidth]{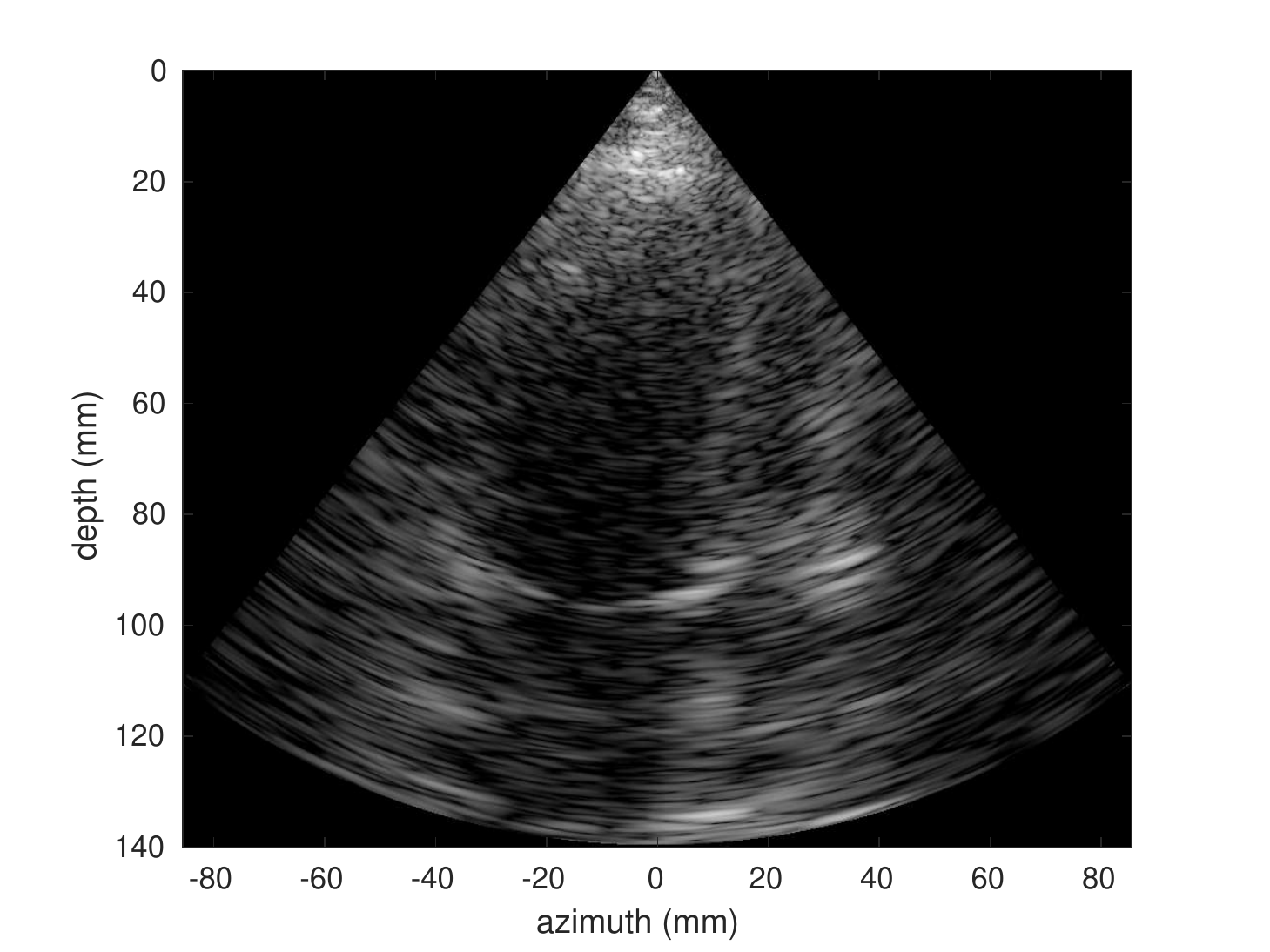}&
		         \\
           &
		\includegraphics[width = 0.33\textwidth]{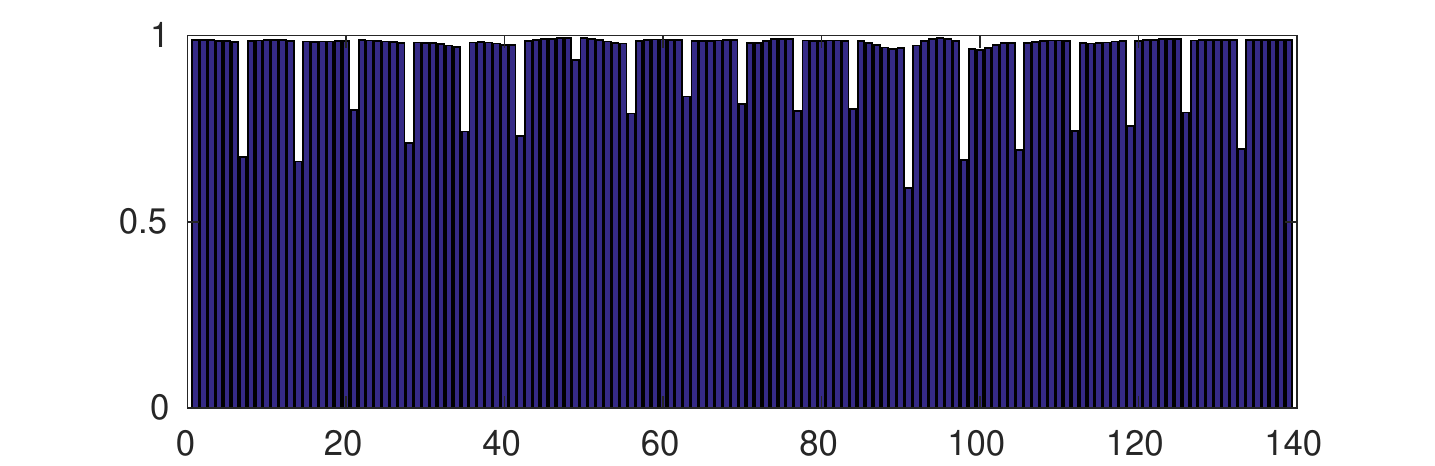} &
		\includegraphics[width = 0.33\textwidth]{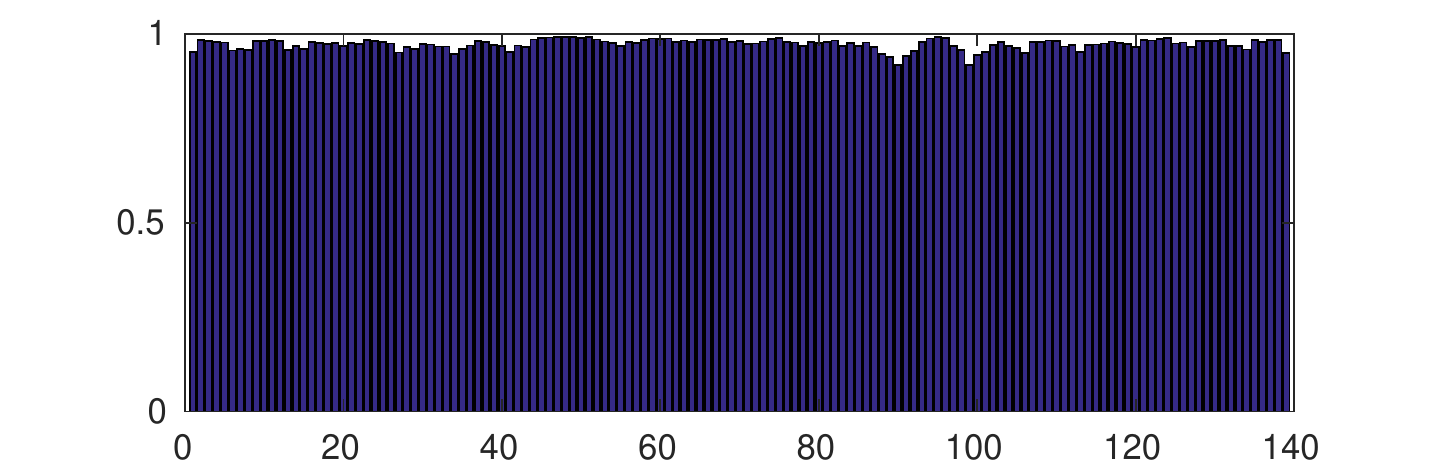}&
                 \\
                & (d) $7-$MLA & (e) Corrected $7-$MLA& \\
	\end{tabular}   \\
   \end{minipage}
    %\smallskip
	\caption{\small \textbf{CNN-based MLA artifact correction tested on cardiac data.}  A test frame from cardiac sequence demonstrating the performance of the proposed artifact correction algorithm. Each image is depicted along with the plot of the correlation coefficients between adjacent lines.}
	\label{CardiacFigs}
\end{figure}

Quantitative results for the cardiac test set are summarized in Table \ref{Table1}. We show a major improvement in decorrelation and SSIM for both $5-$ and $7-$MLA. The corrected $7-$MLA performance approaches that of $5-$MLA, suggesting the feasiblity of larger MLA factors.
Figure \ref{CardiacFigs} shows representative images from each imaging modality. We show that the correlation coefficients profile of the corrected $5-$ and $7-$MLA approaches that of SLA. 

Similarly, quantitative results for the phantom test set are summarized in Table \ref{Table2}, again showing a significant improvement in the image quality for both $5-$ and $7-$MLA. Visual results with the corresponding correlation coefficients profiles are depicted in Figure 1 in the Supplementary Material. 
These results suggest that the networks trained on real cardiac data generalize well to the phantom data without any further training or fine-tuning. For comparison, \cite{bjastad2007impact} reported a decorrelation value of $-1.5$ for a phantom image acquired in a $4-$MLA mode with STB compensation, while we report closer to zero $D_c$ values, $0.457$ for $5-$MLA and $0.956$ for $7-$MLA, which both use a greater decimation rate. The slight dissimilarities in the recovered data can be explained by the acquisition method being used: since the scanned object was undergoing a motion, there is a difference between all but a central line in each MLA group and the matching lines in SLA. We assume that training the network on images of static organs may further improve its performance. Independently, small areas with vertical stripes were observed in several images. In our opinion, the origin of the stripes is a coherent summation of the beamformed lines across the moving object. Since the frame rate of the employed acquisition sequence was slower than of genuine MLA acquisition, the magnitude of this artifact is probably exaggerated.         
\begin{table}
\begin{center}

    \resizebox{.8\textwidth}{!}{\begin{tabular}{|c |c |c |c |c| c |c|} 
 \hline
  &\multicolumn{1}{|c|}{SLA}&\multicolumn{2}{|c|}{$5-$MLA}&\multicolumn{2}{|c|}{$7-$MLA}\\ 
\hline
 & Original  & Original & Corrected & Original & Corrected \\ [0.5ex] 
 \hline
Decorrelation  & $0.03$/$-0.04$  & $22.03$ & $0.69$  & $31.7$ & $0.827$  \\ 
\hline
SSIM  & -   & $0.75$ & $0.876$ & $0.693$ & $0.826$ \\ 
\hline
\end{tabular}
}
\vspace{1mm}
\caption{\small \textbf{Image reconstruction results on cardiac data:} comparison of average decorrelation and SSIM measures between the original and corrected $5-$ and $7-$MLA cardiac images. Decorrelation of SLA is reported in the first column; left and right values in the entry indicate the values calculated for $5-$ and $7-$MLA, respectively.}
\label{Table1}
\end{center}
\end{table}

\begin{figure}[h]
\begin{minipage}[]{\linewidth}
	\begin{tabular}{ c@{\hskip 0.001\textwidth}c@{\hskip 0.001\textwidth}c@{\hskip 0.001\textwidth}c} 

		\includegraphics[width = 0.33\textwidth]{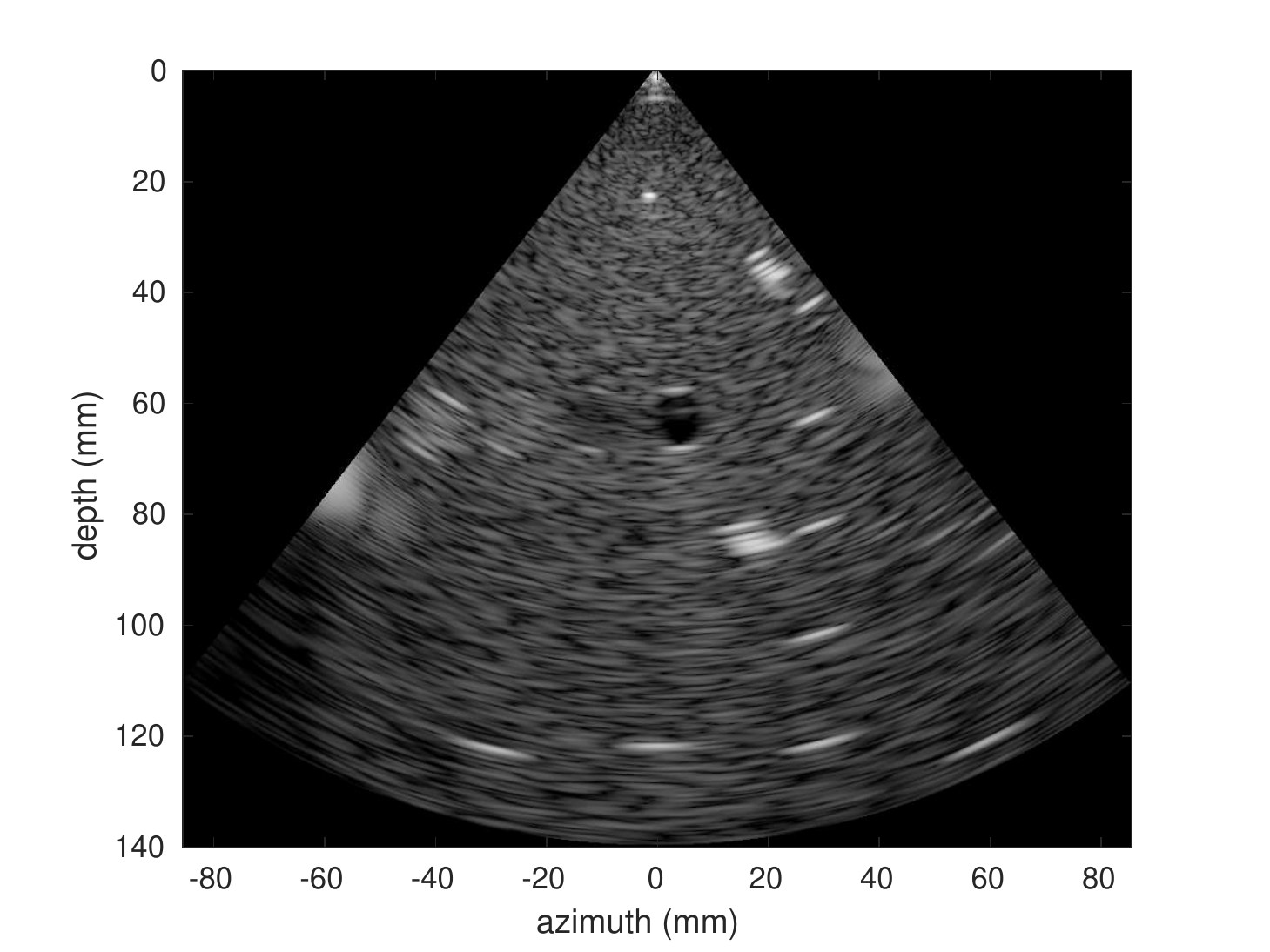} &
		\includegraphics[width = 0.33\textwidth]{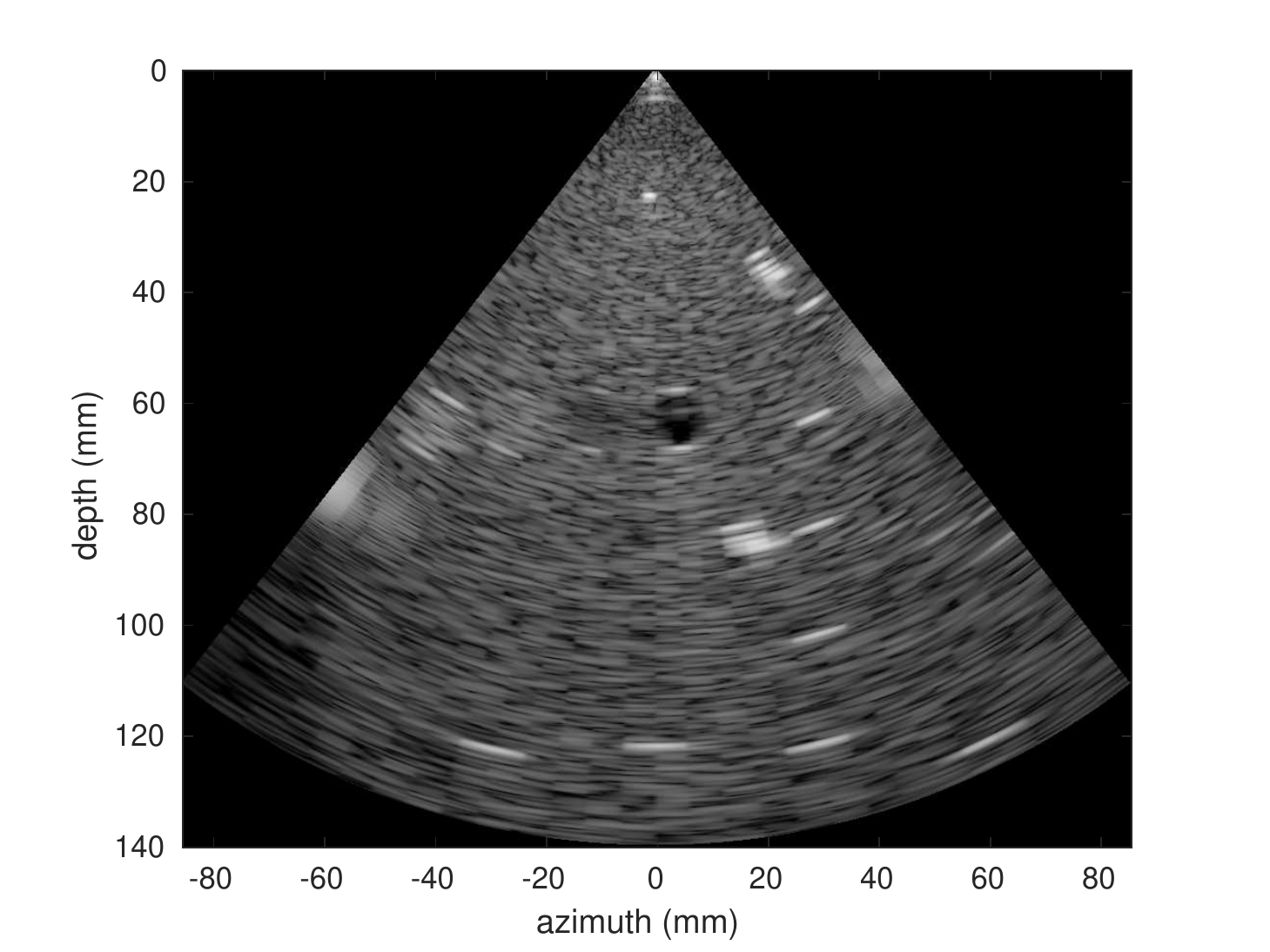} &
		\includegraphics[width = 0.33\textwidth]{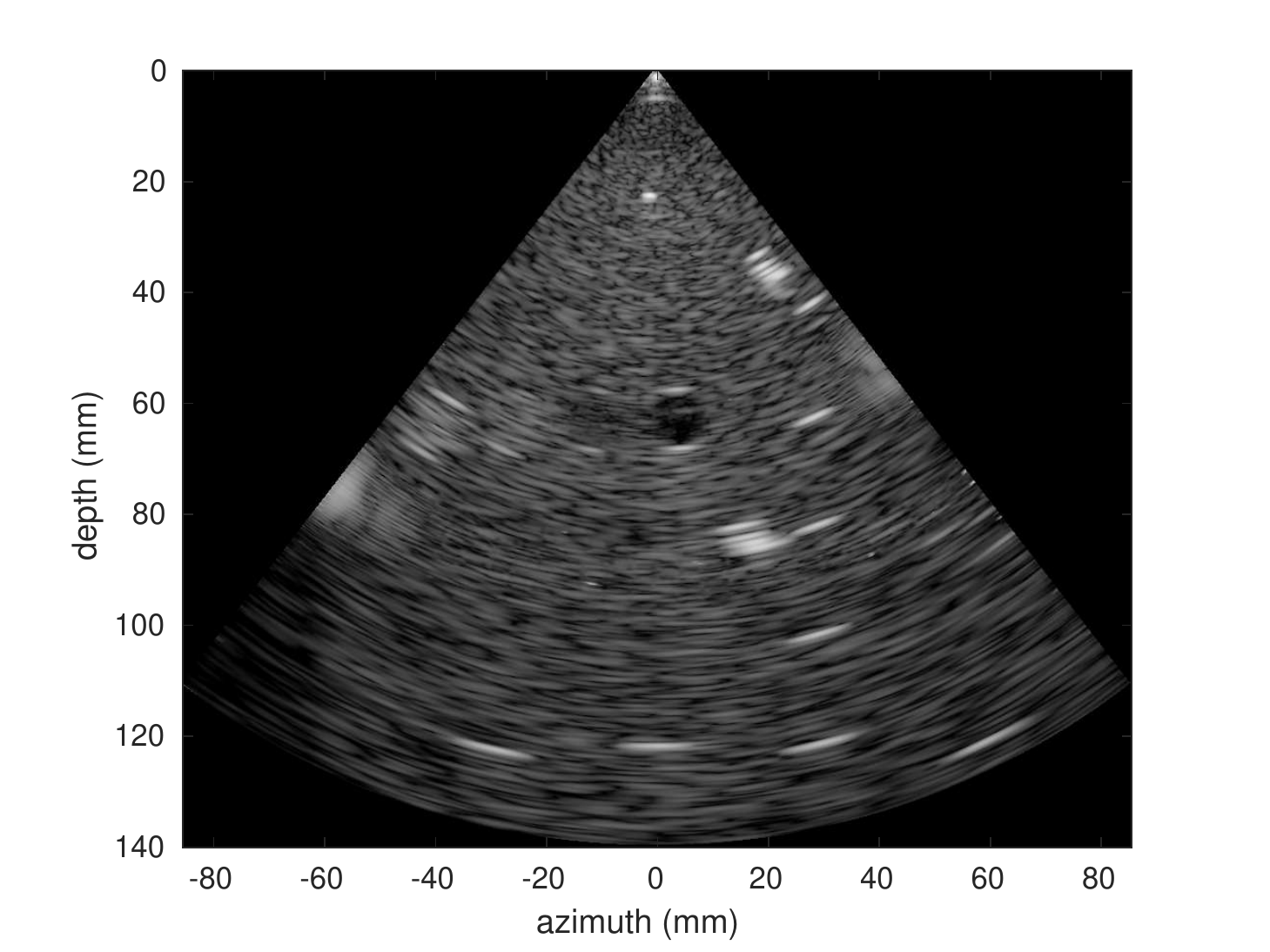} &
	       \\
        \includegraphics[width = 0.33\textwidth]{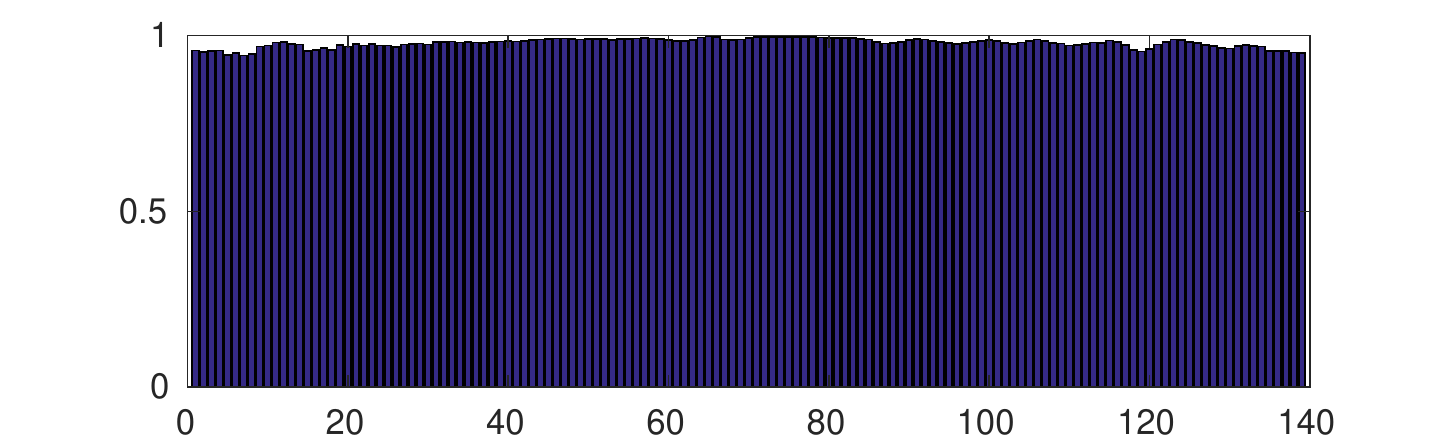} &
		\includegraphics[width = 0.33\textwidth]{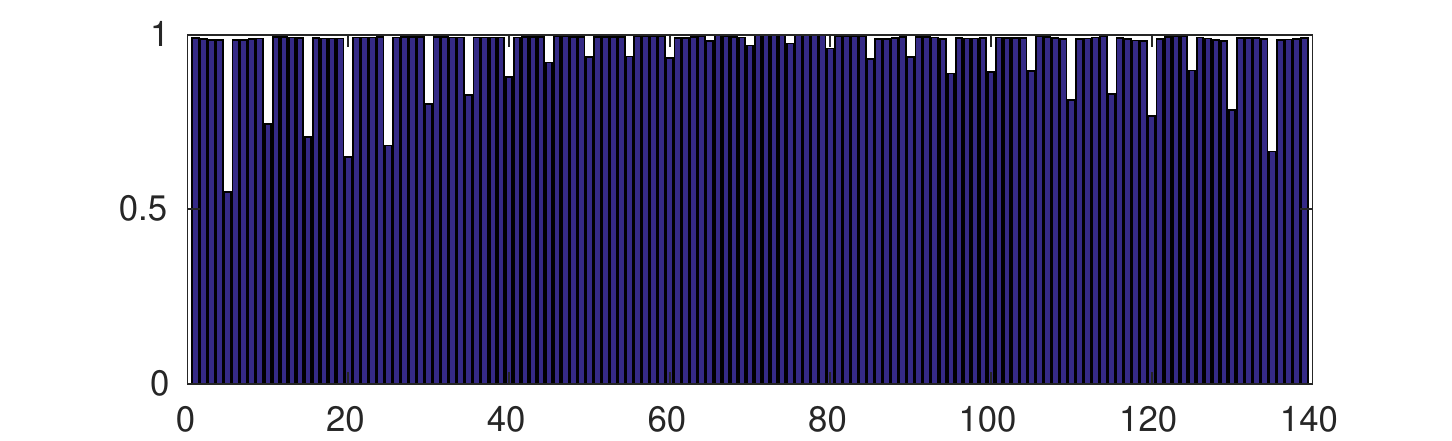} &
		\includegraphics[width = 0.33\textwidth]{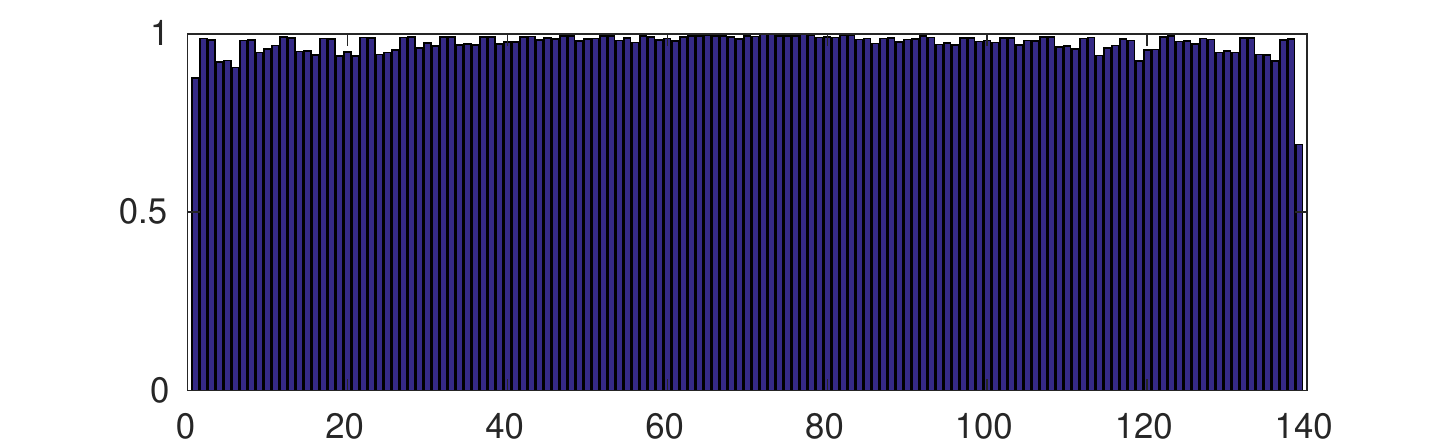}&
         \\
               (a) SLA  & (b) 5MLA & (c) Corrected 5MLA & \\
           &
		\includegraphics[width = 0.33\textwidth]{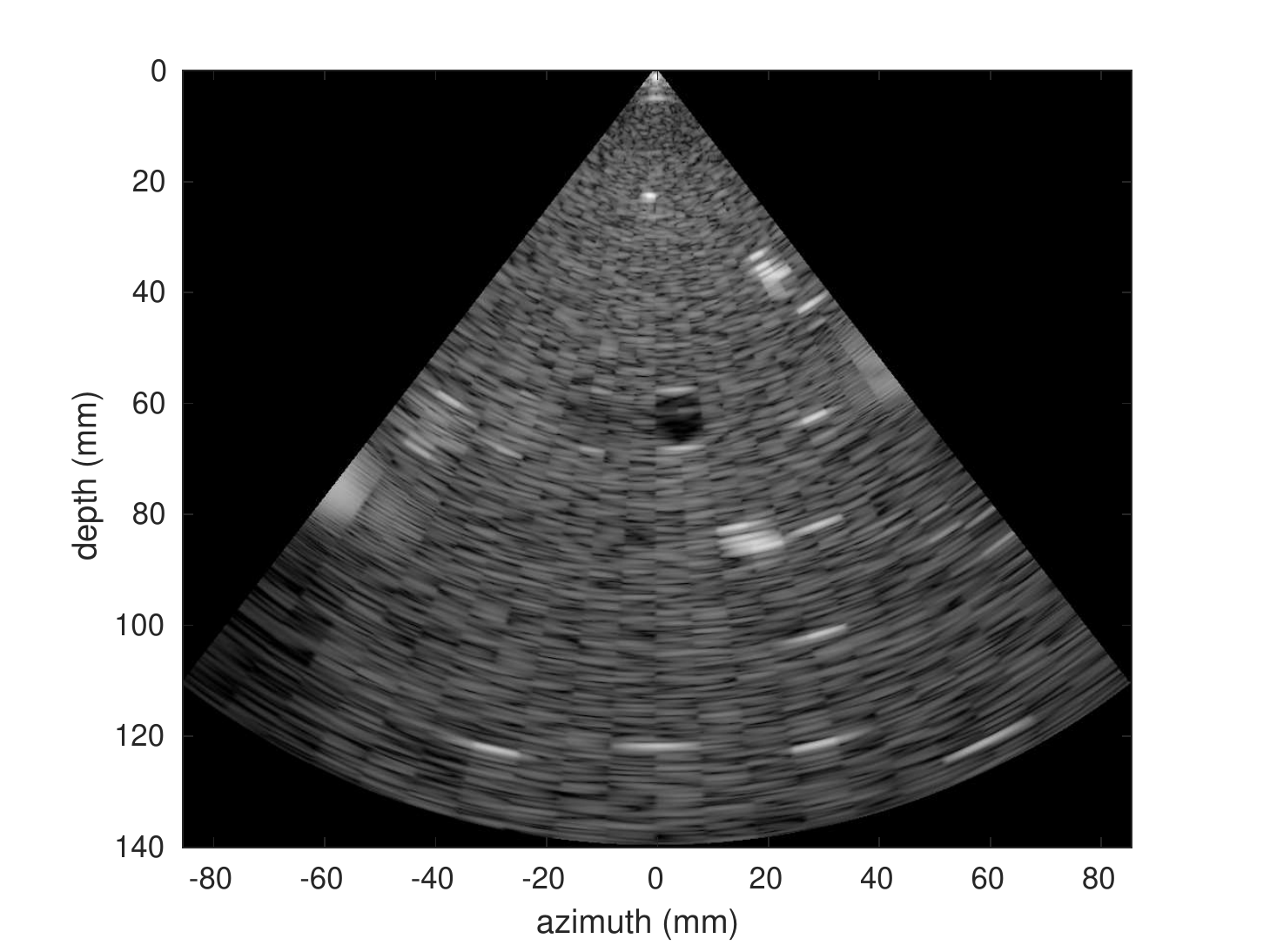} &
		\includegraphics[width = 0.33\textwidth]{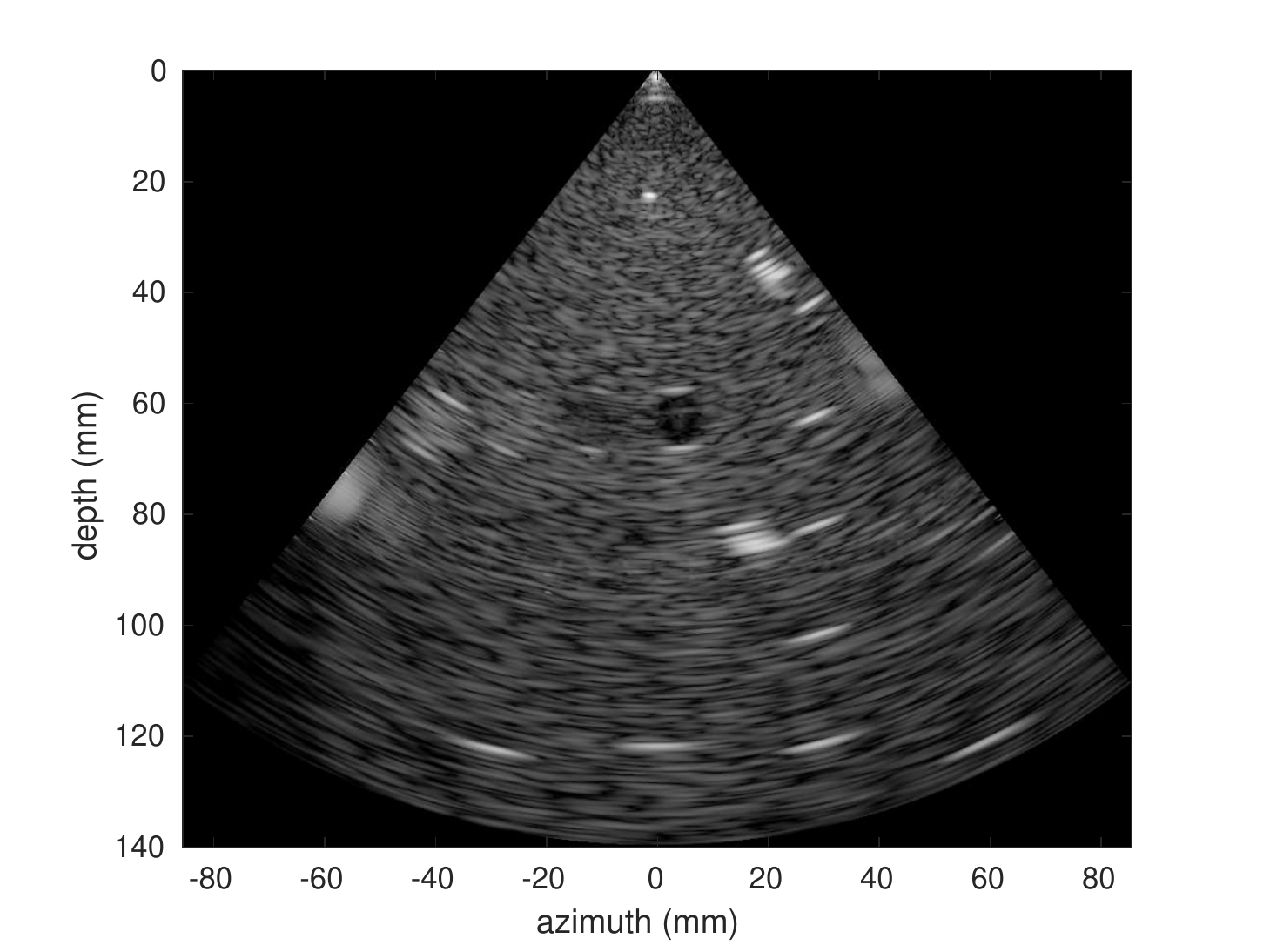}&
		         \\
           &
		\includegraphics[width = 0.33\textwidth]{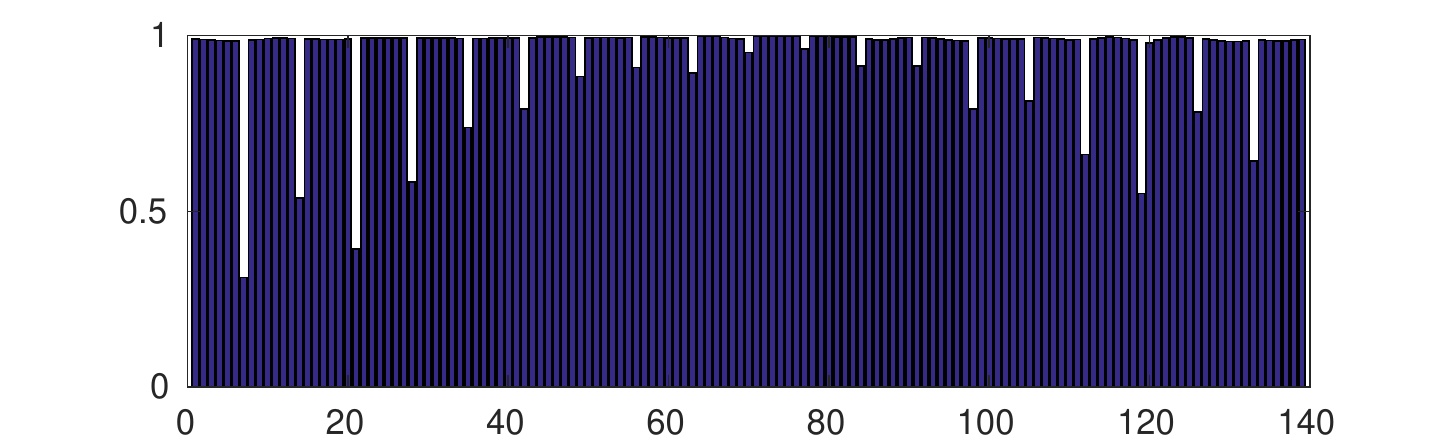} &
		\includegraphics[width = 0.33\textwidth]{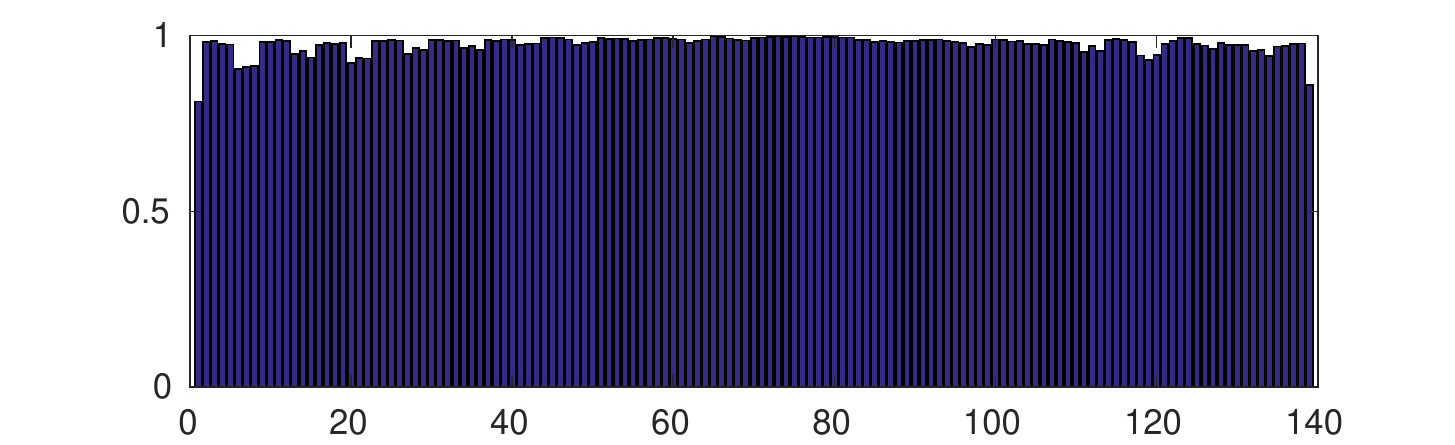}&
                 \\
                & (d) 7MLA & (e) Corrected 7MLA& \\
	\end{tabular}   \\
   \end{minipage}
    %\smallskip
	\caption{\small \textbf{CNN-based MLA artifact correction tested on phantom data} A test frame from the phantom data demonstrating the performance of the proposed artifact correction algorithm. Each image is depicted along with the plot of the cross-correlation coefficient between adjacent lines.}
	\label{PhantomFigs}
\end{figure}

\begin{table}
\begin{center}

    \resizebox{.8\textwidth}{!}{
\begin{tabular}{|c |c |c |c |c| c |c|} 
 \hline
  &\multicolumn{1}{|c|}{SLA}&\multicolumn{2}{|c|}{$5-$MLA}&\multicolumn{2}{|c|}{$7-$MLA}\\ 
\hline
 & Original  & Original & Corrected & Original & Corrected \\ [0.5ex] 
 \hline
Decorrelation  & $0.06$/$-0.089$  & $19.53$ & $0.457$  & $32.34$ & $0.956$  \\ 
\hline
SSIM  & -   & $0.815$ & $0.96$ & $0.793$ & $0.935$ \\ 
\hline
\end{tabular}
}
\vspace{1mm}
\caption{\small \textbf{Image reconstruction results on phantom data:} Comparison of average decorrelation and SSIM measures between original and corrected $5-$ and $7-$MLA phantom images. Decorrelation of SLA is reported in the first column; left and right values in the entry indicate values calculated for $5-$ and $7-$MLA, respectively.}
\label{Table2}
\end{center}
\end{table}

%\begin{figure}[ht]
%\begin{minipage}[]{\linewidth}
%	\begin{tabular}{ c@{\hskip 0.001\textwidth}c@{\hskip 0.001\textwidth}c@{\hskip 0.001\textwidth}c} 
%
%		\includegraphics[width = 0.33\textwidth]{ph1_sla} &
%		\includegraphics[width = 0.33\textwidth]{ph1_5mla} &
%		\includegraphics[width = 0.33\textwidth]{ph1_5mla_corr} &
%	       \\
 %       \includegraphics[width = 0.33\textwidth]{ph1_Cc_sla} &
	%	\includegraphics[width = 0.33\textwidth]{ph1_Cc_5mla} &
	%	\includegraphics[width = 0.33\textwidth]{ph1_Cc_5mla_corr}&
     %    \\
      %         (a) SLA  & (b) $5-$MLA & (c) Corrected $5-$MLA & \\
       %    &
		%\includegraphics[width = 0.33\textwidth]{ph1_7mla} &
		%\includegraphics[width = 0.33\textwidth]{ph1_7mla_corr}&
		 %        \\
          % &
		%\includegraphics[width = 0.33\textwidth]{ph1_Cc_7mla} &
		%\includegraphics[width = 0.33\textwidth]{ph1_Cc_7mla_corr}&
         %        \\
          %      & (d) $7-$MLA & (e) Corrected $7-$MLA& \\
	%\end{tabular}   \\
   %\end{minipage}
    %\smallskip
    %\vspace{-0.2cm}
	%\caption{\small \textbf{CNN-based MLA artifact correction tested on phantom data.} A test frame from the phantom data demonstrating the performance of the proposed artifact correction algorithm is presented. Each image is depicted along with the plot of the cross-correlation coefficient between adjacent lines.}
	%\label{PhantomFigs}
%\end{figure}

\section{Conclusion}
In this paper, we have shown that conventional ultrasound MLA correction can be substituted with an end-to-end CNN performing both optimal interpolation and apodization in order to approximate SLA image quality. In the future, we aim at extending this approach to even earlier stages in multi-line acquisition such as beamforming, assuming it will provide a greater improvement in image quality. Moreover, in a concurrent work \cite{vedula2018high}, we demonstrate that similar method could be applied for other fast US acquisition modalities, such as multi-line transmission (MLT) \cite{mallart1992improved}. 
 
 \section{Acknowledgements}
 This research was partially supported by ERC StG RAPID.
%\afterpage{\clearpage}

\bibliographystyle{splncs}
\bibliography{main}

\end{document}